\documentclass[10pt,twocolumn,letterpaper]{article}

\usepackage{multicol}
\usepackage{multirow}
\usepackage{microtype}      
\usepackage{xcolor}         
\usepackage{graphicx}
\usepackage{colortbl}
\usepackage{makecell}
\usepackage{booktabs}

\definecolor{orange}{RGB}{255,165,0}

\usepackage{cvpr}             

\definecolor{cvprblue}{rgb}{0.21,0.49,0.74}
\usepackage[pagebackref,breaklinks,colorlinks,citecolor=cvprblue]{hyperref}

\def\paperID{3888} 
\def\confName{CVPR}
\def\confYear{2024}

\title{Narrative Action Evaluation with Prompt-Guided Multimodal Interaction}

\author{
    Shiyi Zhang\textsuperscript{1}$^,$\footnotemark[1]~,
    Sule Bai\textsuperscript{1}$^,$\footnotemark[1]~,
    Guangyi Chen\textsuperscript{2},    
    Lei Chen\textsuperscript{3},
    Jiwen Lu\textsuperscript{3},
    Junle Wang\textsuperscript{4},
    Yansong Tang\textsuperscript{1}$^,$\footnotemark[2]~\\
    \textsuperscript{1} Shenzhen International Graduate School, Tsinghua University\\
    \textsuperscript{2} Carnegie Mellon University, Pittsburgh PA, USA\\
    \textsuperscript{3} Department of Automation, Tsinghua University\ \ \ 
    \textsuperscript{4} Tencent\\
    {\tt \small \{sy-zhang23@mails.,bsl23@mails.,tang.yansong@sz.\}tsinghua.edu.cn}\\
}

\begin{document}

\makeatletter

\let\@oldmaketitle\@maketitle
\renewcommand{\@maketitle}{\@oldmaketitle
\begin{minipage}{\textwidth}
\vspace{-0.8cm}
\centering
\includegraphics[width=1\linewidth]{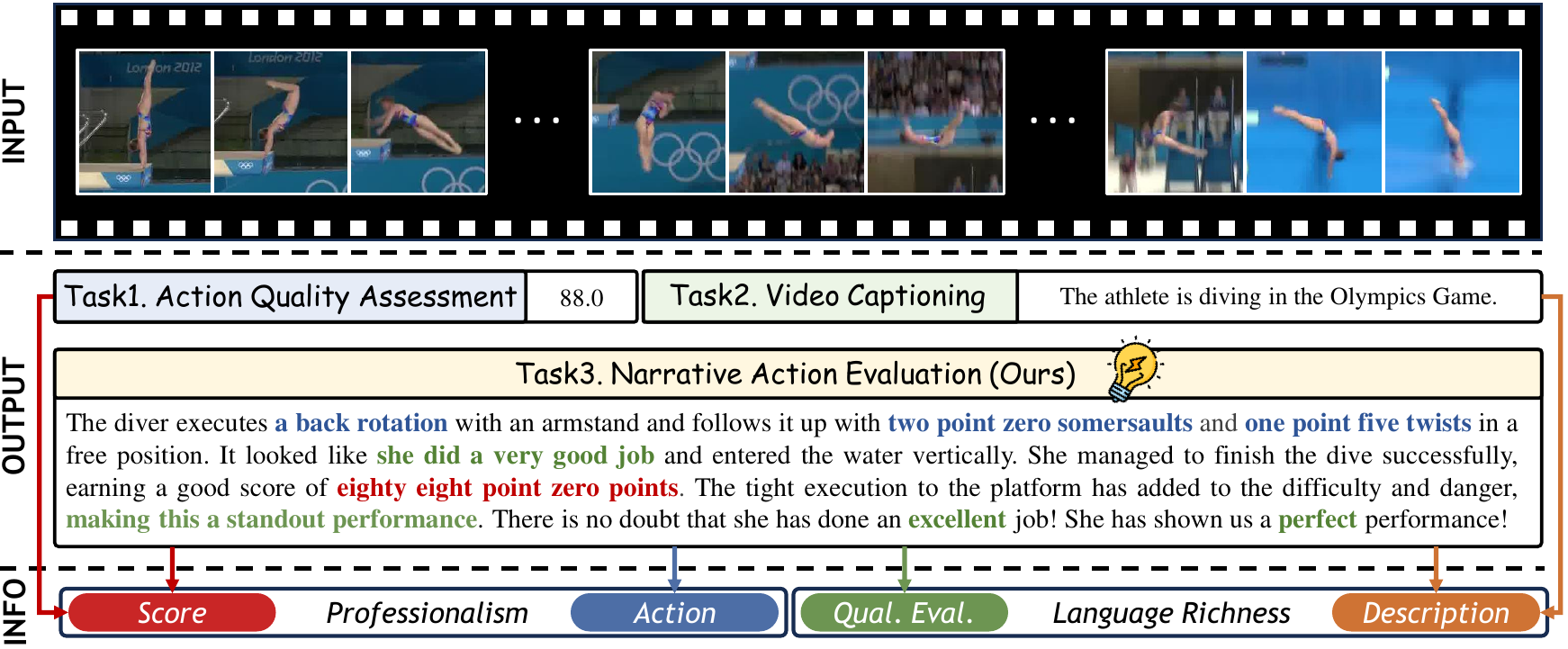}
    \vspace{-6.5mm}
    \captionof{figure}{A comparison of our proposed narrative action evaluation (NAE) task with action quality assessment (AQA) and video captioning. The three lines in the figure represent the input video, the outputs of the three tasks, and the information contained in each task. In comparison to AQA, NAE provides rich language \textcolor[RGB]{207,115,52}{\textbf{\underline{\textit{descriptions}}}}. When compared to Video Captioning, NAE includes much more evaluation information such as \textcolor[RGB]{192,0,0}{\textbf{\underline{\textit{scores}}}}, \textcolor[RGB]{47,85,151}{\textbf{\underline{\textit{actions}}}}, and \textcolor[RGB]{109,149,82}{\textbf{\underline{\textit{qualitative evaluations}}}}, which is often rigorous and granular. In general, NAE aims to strike a balance between the professionalism of assessment information and the richness of language. This duality is both the characteristic and challenge of NAE.}
    \label{fig:overview}
    \vspace{+2mm}
\end{minipage}}
\makeatother

\maketitle
\footnotetext[1]{Equal contribution}
\footnotetext[2]{Corresponding author}

\begin{abstract}
In this paper, we investigate a new problem called narrative action evaluation (NAE). NAE aims to generate professional commentary that evaluates the execution of an action. Unlike traditional tasks such as score-based action quality assessment and video captioning involving superficial sentences, NAE focuses on creating detailed narratives in natural language. These narratives provide intricate descriptions of actions along with objective evaluations. NAE is a more challenging task because it requires both narrative flexibility and evaluation rigor. One existing possible solution is to use multi-task learning, where narrative language and evaluative information are predicted separately. However, this approach results in reduced performance for individual tasks because of variations between tasks and differences in modality between language information and evaluation information. To address this, we propose a prompt-guided multimodal interaction framework. This framework utilizes a pair of transformers to facilitate the interaction between different modalities of information. It also uses prompts to transform the score regression task into a video-text matching task, thus enabling task interactivity. To support further research in this field, we re-annotate the MTL-AQA and FineGym datasets with high-quality and comprehensive action narration. Additionally, we establish benchmarks for NAE. Extensive experiment results prove that our method outperforms separate learning methods and naive multi-task learning methods. Data and code are released at \href{https://github.com/shiyi-zh0408/NAE_CVPR2024 }{here}.

\end{abstract}    
\vspace{-6pt}
\section{Introduction}
\label{sec:intro}

\vspace{-6pt}
Recent years have witnessed substantial advancements in action description methods, including video captioning\cite{lin2022swinbert, luo2020univl, yamazaki2022vlcap, yamazaki2022vltint, livalue2021, liu2020sibnet, zhang2021open} that describes content (Task2 in Figure \ref{fig:overview}) and quality assessment\cite{bertasius2017baller, tang2020uncertainty, yu2021group, shiyi2023logo, xu2022finediving, parmar2019and, parmar2019and, venkataraman2015dynamical, parmar2017learning} that evaluates quality (Task1 in Figure \ref{fig:overview}). However, most of the works only provide a depiction of actions from a singular perspective. How to balance information from multiple dimensions and provide a more comprehensive, rich, and professional evaluation remains blank. As shown in Figure \ref{fig:overview}, we introduce the Narrative Action Evaluation (NAE) task which aims to utilize narrative language to holistically evaluate actions from multiple perspectives, such as action depictions, professional objective evaluations, and qualitative evaluations. This approach merges the language richness of narrative description with the professionalism of expert analysis, thereby providing a multifaceted and thorough evaluation. The NAE presents numerous practical applications in the real world.
For example, these narrative assessments could serve as an AI commentator providing professional descriptions of sports events, or as a fitness coach offering real-time feedback in natural language for physical actions.

As shown in Figure \ref{fig:overview}, the task of Narrative Action Evaluation is notably complex as it necessitates a delicate equilibrium between narrative flexibility and evaluation precision. Striking this balance is a substantial challenge as these two objectives sometimes conflict with each other. To be concrete, Figure \ref{fig:overview} shows that some evaluation information (such as scores and actions) that require very high accuracy often only occupies a small proportion of the output sentence. However, language models often pursue diversity in the generation, which contradicts the rigor of evaluation information. To resolve this contradiction, models should be guided to focus on the evaluation information during text generation. An intuitive solution is using multi-task learning, which generates the text and predicts evaluation information at the same time. Similarly, \cite{parmar2019and} utilized a multi-task learning paradigm in which tasks are parallel to each other. Through the relatively independent training process of three tasks, the backbone shared by the tasks was refined. However, experiments have shown that such a paradigm improves every single task very little and even weakens the performance in some cases. The reason for this phenomenon is that such a multi-task learning paradigm, which only trains multiple tasks in parallel, may ultimately lead to a lack of interaction between features that focus on different tasks or even different modalities. Finally, it may confuse the model when balancing multiple tasks or even lose sight of one another.

In order to solve this problem, we propose a new framework, Prompt-Guided Multimodal Interaction, to encourage interactions between different modalities and tasks to aid the joint learning of description and evaluation. Specifically, we perform the first multimodal interaction by augmenting the learnable prompts which contain score information with video features through Context-Aware Prompt Learning. Then, in Score-Guided Tokens Learning, we formulate the score prediction as a video-text matching task, letting the video features perceive the score information of the language modality, and perform the second multimodal interaction under the guidance of the prompt. This process establishes the interaction between the score prediction task and the text generation task. Afterward, we combine the obtained multimodal embeddings with a learnable template, which contains the predicted score, action information, and the learnable prompt, as the input to the text decoder. Finally, we use the Tri-Token Attention Mask to guide the decoder to focus on professional evaluation information and video information in the input, ultimately generating the narrative evaluations.

Additionally, we find that the existing datasets are insufficient for supporting research in NAE. For instance, text labels in MTL-AQA\cite{parmar2019and}, transcribed directly from video speeches, are too colloquial, incoherent, and noisy to be effectively used as texts for NAE. To propel further research in NAE, we re-annotate the MTL-AQA and FineGym\cite{shao2020finegym} datasets as shown in Figure \ref{fig:textreconstruction}, ensuring high-quality and comprehensive action narration. We will make our code and data publicly available to support further progress on the NAE task. Moreover, we establish benchmarks using these datasets, demonstrating that our framework significantly outperforms the approaches based on previous state of the arts.


The contributions of this paper can be summarized as: (1) We propose a new task, Narrative Action Evaluation, which aims to generate professional commentary to evaluate action execution. And we re-annotate MTL-AQA and FineGym datasets for this task. (2) To tackle NAE, we propose a new framework, Prompt-Guided Multimodal Interaction, which uses prompts to integrate information from different modalities, realizing better mutual promotion between multiple tasks. (3) Experiments show that our framework surpasses the baselines based on previous state-of-the-art approaches.

\vspace{-7pt}
\section{Related Work}
\label{sec:relatedwork}
\vspace{-5pt}


\textbf{Action Quality Assessment.} AQA is a task of evaluating the quality of actions performed in videos across various domains such as sports events\cite{gordon1995automated, parmar2019action, jug2003trajectory, pervse2007automatic, pirsiavash2014assessing, parmar2019and, venkataraman2015dynamical, parmar2017learning, shiyi2023logo}, healthcare\cite{malpani2014pairwise, zhang2014relative, sharma2014video, zhang2011video, zia2015automated, zia2018video}, and others. Most existing approaches\cite{doughty2019pros,tang2020uncertainty, yu2021group,gordon1995automated, pirsiavash2014assessing, venkataraman2015dynamical, parmar2017learning,malpani2014pairwise, sharma2014video} treat AQA as a regression task, employing diverse video representations as input and training the model with scores in a supervised manner. For example, Xu \textit{et.al.} \cite{xu2022finediving} utilizes fine-grained action information to assist in score prediction. Bai \textit{et.al.} \cite{bai2022action} constructs learnable action queries to encode action information in videos. While the regression paradigm has shown impressive performance in predicting precise scores, it falls short in providing comprehensive evaluations due to the single score output. Parmar \textit{et.al.} \cite{parmar2019and} tackles the issue by proposing a new dataset with caption labels and reformulating AQA as a multi-task parallel learning paradigm. Nonetheless, the captions within the dataset are considerably noisy and informal, thus limiting the model's capacity to generate professional commentary. To solve this, we re-annotate the captions and design a new framework to enable the multimodal information interaction guided by prompts.
\vspace{-12pt}
\paragraph{Video Captioning.} Video Captioning\cite{pei2019memory, chen2019deep, livalue2021} is a task of generating language descriptions of the video content. The most common method for video captioning is based on the encoder-decoder architecture. Several works\cite{aafaq2019spatio,shi2020learning, pan2020spatio, luo2020univl, zhang2021open, yamazaki2022vlcap, yamazaki2022vltint} propose to use different vision encoders\cite{he2016deep, hara2018can, feichtenhofer2019slowfast, xie2018rethinking, Szegedyinception} to extract features from video frames, and a language decoder to generate captions using the visual features. Recently, SwinBERT\cite{lin2022swinbert} utilizes Video Swin Transformer\cite{liu2022video} as the vision encoder and a Transformer-based module\cite{DevlinCLT19BERT} as the text decoder, resulting in an end-to-end model which directly takes video frames as input. While video captioning models can effectively capture and describe visual content, they struggle to provide detailed and reliable assessments of action quality. In this paper, we propose a prompt-guided multimodal interaction multi-task learning approach, which can precisely depict the video content and narratively evaluate the quality of actions.

\vspace{-7pt}

\section{Dataset}
\label{sec:dataset}
\vspace{-5pt}

To facilitate research on the NAE task, we construct new video-text pair datasets. Specifically, we re-annotate MTL-AQA \cite{parmar2019and} and FineGym \cite{shao2020finegym} datasets with rich narrative texts for videos, including multidimensional evaluation information such as scores, actions, and qualitative evaluations. 

\begin{figure}[t]
    \centering
    \includegraphics[width=\linewidth]{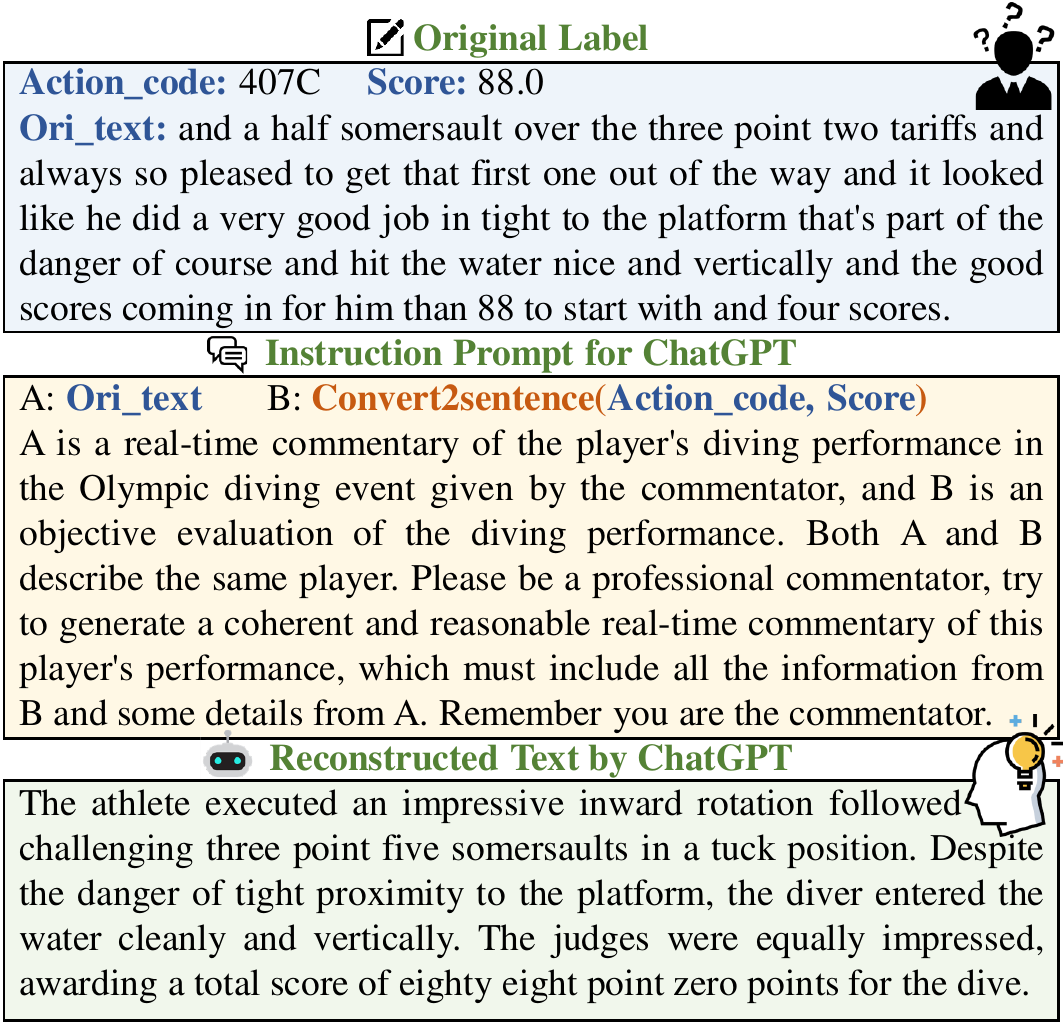}
    \vspace{-20pt}
    \caption{The process of re-annotating a sample using ChatGPT, based on existing action and score labels. \textit{Convert2sentence(act, score)} constructs a pre-fixed template to insert the action and score information into the template to generate a complete sentence.}
    \label{fig:textreconstruction}
    \vspace{-10pt}
\end{figure}

As shown in Figure \ref{fig:textreconstruction}, we take MTL-AQA as an example to explain our re-annotation process. The MTL-AQA dataset is a multi-task AQA dataset that provides three types of annotations for each video: action codes, scores, and text transcribed from video audio (an example is noted as \textcolor[RGB]{109,149,82}{\textbf{Original Label}} in Figure \ref{fig:textreconstruction}). Among them, the action codes reflect all fine-grained professional action information performed by athletes. Although MTL-AQA already includes video-text pairs, the quality of its text labels is often poor because they are directly transcribed from the audio of the videos and contain a lot of interference information (an example is noted as \textcolor[RGB]{47,85,151}{\textbf{Ori\_text}} in Figure \ref{fig:textreconstruction}). In addition, the commentary information in the videos often only contains subjective evaluations from commentators and tone words; there is generally a lack of professional rigorous evaluation information such as action types and scores in the text. To solve this problem, we integrate and reconstruct the textual information in MTL-AQA with ChatGPT \cite{instructgpt} using prompts. The prompts as shown in Figure \ref{fig:textreconstruction} instruct ChatGPT to generate texts including professional evaluation information while preserving details from the original transcription captions. Additionally, to make generated texts more diverse, we design five versions of prompts for both datasets so that ChatGPT can generate five different evaluative texts for each video (an example of the reconstructed text is shown in Figure \ref{fig:textreconstruction}). Table \ref{tab:datasetcomparison} shows experiments using video captioning methods on the datasets before and after our re-annotation, the significantly better results also validate the effectiveness of our re-annotation. 

FineGym offers fine-grained action information for each video. During the re-annotation, we first annotate scores and transcription captions for each sample (taken from the original competition videos). Then following a similar approach mentioned above with MTL-AQA, we insert the actions, scores, and transcription captions into prompts that are inputted into ChatGPT to generate narrative evaluative texts. 

\begin{table}
    \caption{Comparison of video captioning results using video captioning methods on the datasets before and after our re-annotation.}
    \vspace{-8pt}
    \centering
    \fontsize{8}{14}\selectfont \renewcommand{\arraystretch}{0.6}
    \setlength{\tabcolsep}{2.7mm}{ 
    \begin{tabular}{l|ccc|ccc}
    \toprule
         \multirow{2}{*}{Method}&  \multicolumn{3}{c|}{MTL-AQA}&  \multicolumn{3}{c}{FineGym}\\
         \cmidrule(lr){2-7} & B4& M& C& B4& M&C\\
         \midrule
         \multicolumn{7}{l}{\emph{Trained with the original narration}} \\
         \midrule
         VLTinT\cite{yamazaki2022vltint}   &  2.5&  12.4&  6.1&  2.1&  10.5&  5.7\\
         SwinBert\cite{lin2022swinbert} &  3.3&  12.3&  7.3&  3.0&   11.9&  6.6\\
         \midrule
         \multicolumn{7}{l}{\emph{Trained with our re-annotated narration (Ours)}} \\
         \midrule
         VLTinT\cite{yamazaki2022vltint}   &  22.5 &  19.9 &  14.4 &  15.4 &  17.6 & 14.2\\
         SwinBert\cite{lin2022swinbert} &  40.2 &  26.4 &  16.2 &  27.4 &  21.0 & 17.8\\
         \bottomrule
    \end{tabular}}
    \vspace{-18pt}
    \label{tab:datasetcomparison}
\end{table}

Finally, to ensure the accuracy of the evaluation information in the generated texts, we hire 8 professional divers and gymnasts to check the generated video-text pairs. Each pair is checked by two athletes to ensure that there are no changes made to the action and score information in the generated text and that the correspondence between scores and quantitative evaluations in the text is correct. For example, when a player scores 98 points, the quantitative evaluation in the text should be positive rather than negative. If there are any errors in a sample's text, we ask ChatGPT to iterate again until it passes the inspection. The entire annotation and checking process above takes about 150 hours to complete. For more details, please refer to supplementary materials.
\begin{figure*}[t]
    \centering
    \includegraphics[width=\linewidth]{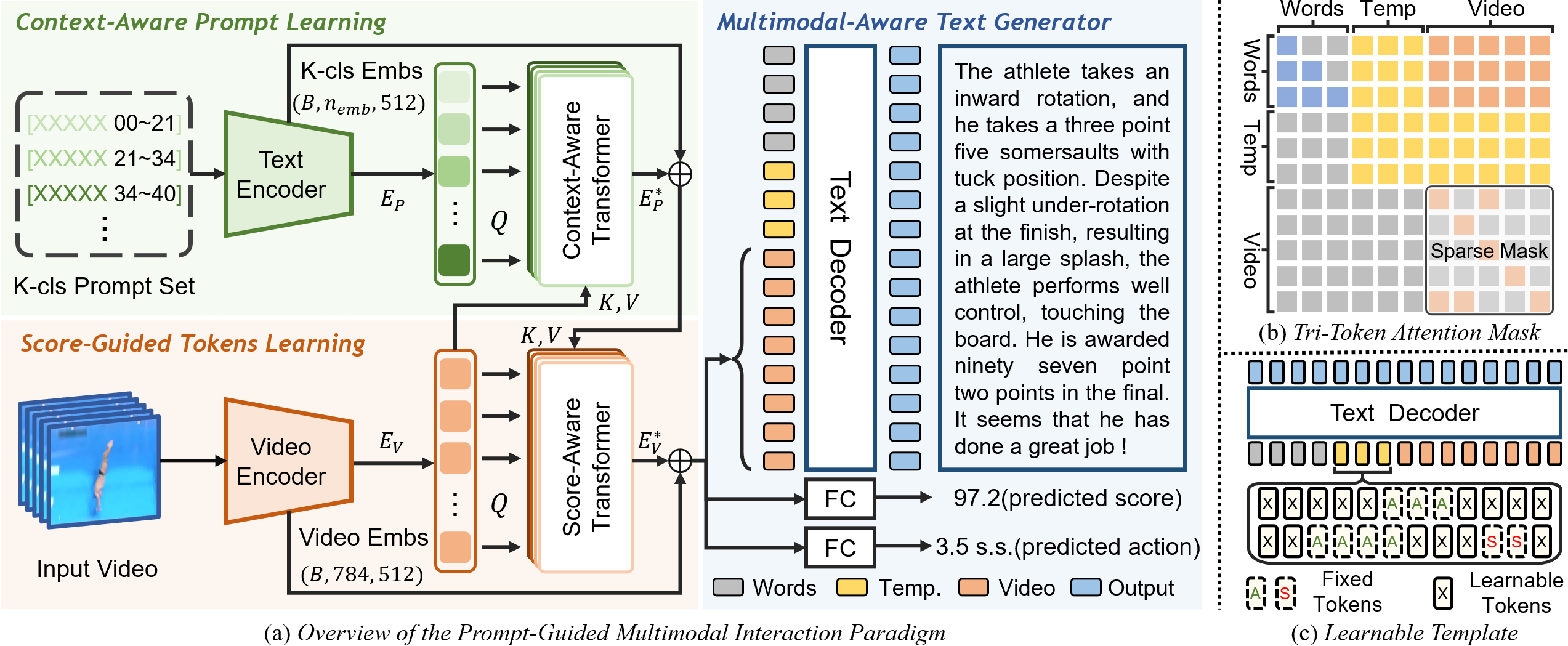}
    \vspace{-20pt}
    \caption{The left part shows an overview of our Prompt-Guided Multimodal Interaction paradigm. First, we send the K-class Prompts into the text encoder to get K-class Prompt Embeddings. After that, we perform Context-Aware Prompt Learning using the video features based on Context-Aware Transformer. Second, in Score-Guided Tokens Learning, we interact the video embeddings from the video encoder with the K-class Prompts mentioned above through Score-Aware Transformer. Thirdly, we utilize Multimodal-Aware Text Generator with the Tri-Token Attention Mask to integrate the multimodal tokens from Score-Guided Tokens Learning and generate the text. The upper right part shows the Tri-Token Attention Mask and the bottom right part shows the learnable template in Multimodal-Aware Text Generator.}
    \label{fig:framework}
    \vspace{-10pt}
\end{figure*}
\section{Approach}
\label{sec:approach}

\subsection{Overview}
\label{subsec:overviewpipline}

An overview of our Prompt-Guided Multimodal Interaction is shown in the left part of Figure \ref{fig:framework}. Our framework consists of three parts. Firstly, in Context-Aware Prompt Learning, we employ a cross-attention module to refine the learnable prompt embeddings $E_P$, which contain score information, with contextual features from video embeddings $E_V$. During this stage, $E_V$ acts as the prompt to guide $E_P$ to perceive contextual information, thus fusing the language modality and video modality. This process can be represented as:
\begin{equation}
    \setlength{\abovedisplayskip}{2pt}
    \setlength{\belowdisplayskip}{2pt}
    E_P^*=\textbf{MHCA}(E_P,E_V)+\gamma_1E_P,
    \label{eq:caplMHCA}
\end{equation} where $\textbf{MHCA}(,)$ indicates multi-head cross-attention whose first parameter denotes ``\textit{query}'', and the second parameter means ``\textit{key}'' and ``\textit{value}''; $E_P^*$ means refined context-aware prompt embeddings; and $\gamma_1$ indicates learnable coefficient.


Secondly, in Score-Guided Tokens Learning, we adopt a similar approach to Context-Aware Prompt Learning. Specifically, we use the refined score-aware embeddings $E_P^*$ mentioned above as the prompt to guide video embeddings $E_V$ to perceive the score information, thus achieving the second multimodal interaction. This process can be represented as:
\begin{equation}
    \setlength{\abovedisplayskip}{2pt}
    \setlength{\belowdisplayskip}{2pt}
    E_V^*=\textbf{MHCA}(E_V,E_P^*)+\gamma_2E_V,
    \label{eq:pgtlMHCA}
\end{equation} where $E_V^*$ denotes refined video embeddings integrated with score-aware embeddings; $\gamma_2$ denotes learnable coefficient. 

Finally, we use Multimodal-Aware Text Generator to generate the narrative evaluation in an autoregressive paradigm, whose input consists of word tokens already produced, learnable template tokens (bottom right of Figure \ref{fig:framework}), and refined video embeddings from Score-Guided Tokens Learning. The learnable template tokens contain score information and action information which is predicted by passing the refined video embeddings $E_P^*$ through an MLP respectively. During the generation of narrative texts, we use the Tri-Token Mask (upper right of Figure \ref{fig:framework}) to guide the text generation process to focus on professional evaluation information from the learnable template and video information from the refined video embeddings. This process can be represented as:
\begin{equation}
    \setlength{\abovedisplayskip}{3pt}
    \setlength{\belowdisplayskip}{3pt}
    Input=\textbf{Concat}(Word_{1:i-1};Template;E_V^*),
\end{equation}
\begin{equation}
    \setlength{\abovedisplayskip}{3pt}
    \setlength{\belowdisplayskip}{3pt}
    Word_{i}=\textbf{Decoder}[\textbf{Mask}_{\textbf{TT}}(Input)],
\end{equation} where $Word_i$ indicates the $i$-th word during generation, $\textbf{Concat}(;;)$ denotes concatenation operation, $\textbf{Mask}_{\textbf{TT}}$ represents Tri-Token Mask. We will now introduce the three parts in our framework separately in the following texts.

\subsection{Context-Aware Prompt Learning}
\label{subsec:contextawarepromptlearning}

\paragraph{Textual Prompt Construction.} To guide the model in generating texts with accurate evaluation information, we incorporate the score into the generation process. One simple approach is to utilize the video features acquired from a pre-trained video backbone and simultaneously conduct score prediction and text generation using separate heads. However, this approach overlooks the distinctions between the score regression task and the text generation task. Furthermore, scores, actions, videos, and texts belong to distinct modalities; nevertheless, this approach fails to account for these modality differences. To tackle this, we reformulate the score prediction problem as a video-text matching problem. Specifically, we employ the learnable prompt to convert numerical scores into textual form. To represent all possible scoring situations with a limited number of categories in the video-text matching problem, we cannot directly use text to represent specific scores because scores are continuous and there are infinitely many of them. Therefore, we use text to represent finite intervals of scores. When predicting scores, different score intervals represent different categories. These intervals do not overlap, and the beginning and end of two adjacent score intervals are the same. To this end, we obtain multiple score intervals with similar distribution probabilities in the sample space. Concretely, we first collect a list of all scores $\textbf{S}=[S_1,..., S_N]$ from training samples. Then, we sort the list in ascending order to obtain $\textbf{S}^*=[S^*_1,..., S^*_N]$. Given the interval numbers R, the partitioning algorithm gives the bounds of each interval $\mathcal{I}^r=(\zeta^r_{left},\zeta^r_{right})$ as:
\begin{equation}
    \setlength{\abovedisplayskip}{3pt}
    \setlength{\belowdisplayskip}{3pt}
    \zeta^r_{left}=\textbf{S}^*(\lfloor(N-1) \times \frac{(r-1)}{R}\rfloor),
\end{equation}
\begin{equation}
    \setlength{\abovedisplayskip}{3pt}
    \setlength{\belowdisplayskip}{3pt}
    \zeta^r_{right}=\textbf{S}^*(\lfloor(N-1) \times \frac{r}{R}\rfloor), 
\end{equation}where $\textbf{S}^*(i)$ represents the $i$-th element of $\textbf{S}^*$. 
{\parfillskip=0pt
Then, we express the score intervals of different categories in textual form, and connect a learnable prompt of a certain length in front, so as to obtain the K-class score-aware textual prompt. For example, the score range from $25.6$ to $36.3$ can be expressed as ``[XXXXXX] twenty-five point six to thirty-six point three'', where ``[X]'' is a learnable token. Until then, we obtain the K-class prompt set with score information. }
\paragraph{Context-Aware Prompting.} After constructing prompts with score information, our model augments the K-class prompt embeddings from the text encoder with contextual information from video embeddings. Specifically, we use a transformer decoder to utilize the video features from the video encoder as ``\textit{key}'' and ``\textit{value}'', and the K-class prompt embeddings from the text encoder as ``\textit{query}'', to refine the prompt embeddings with context information from the video, so as to integrate prompt embeddings with context information. Then these prompt embeddings will be used in the video-text matching during Score-Guided Tokens Learning.
\subsection{Score-Guided Tokens Learning} 
\label{subsec:learnabletemplatetokens}
In \ref{subsec:contextawarepromptlearning}, we have obtained score-aware prompt embeddings that have perceived video context information. In Score-Guided Tokens Learning, we use a cross-attention module called Score-Aware Transformer, which is symmetrical to Context-Aware Transformer in \ref{subsec:contextawarepromptlearning}, to integrate video embeddings from the video encoder with prompt embeddings. 

In the Score-Aware Transformer, we aim to enhance the attention of input video information toward score-aware prompt embeddings that correspond to the video. This ensures the accurate integration of score information into the video features. As mentioned above, we use score intervals to convert the score prediction task into a classification task. So we supervise the input video to focus more on its corresponding score interval using cross-entropy loss, which is represented as, $\mathcal{L}_{CE_S}=-\sum_i p_ilog\hat{p}_i$, where $p_i$ indicates the possibility that the predicted score belongs to the $i$-th interval defined in \ref{subsec:contextawarepromptlearning}. This process completes the filter of text-based score information and refines video information.

Then, we obtain the video tokens that incorporate score information. We merge these video tokens with word tokens and template tokens, then input them into the text decoder for text generation. Details of the template tokens will be explained in \ref{subsec:multimodeltextgenerator}. Additionally, we pass the video tokens through two heads that predict the score and action respectively. The score head is an MLP. The action head consists of multiple MLPs, and each MLP corresponds to different parts' actions in videos. The MSE loss $\mathcal{L}_{MSE}$ and CE loss $\mathcal{L}_{CE_A}$ supervises the score and action prediction respectively.

\subsection{Multimodal-Aware Text Generator}
\label{subsec:multimodeltextgenerator}
{\parfillskip=0pt
Following \cite{lin2022swinbert}, we use a Transformer-based generator as the text decoder to generate the natural language description. As shown in the bottom right of Figure \ref{fig:framework}, the text decoder has input from multiple modalities, which include word tokens already generated, score-aware video tokens from Score-Guided Tokens Learning, and learnable template tokens. This template includes three parts: learnable tokens, unlearnable action, and score tokens. Specifically, in \ref{subsec:learnabletemplatetokens}, we use two heads to predict the scores and the fine-grained action categories for each part in videos respectively. Therefore, we can obtain a score, and action types correspond to different parts of videos. We convert them into textual form and insert them as fixed tokens into the template. Learnable tokens are inserted between these fixed tokens, concatenating action, and score information into a complete sentence.}

Moreover, we use the Tri-Token Attention Mask (as shown in the upper right of Figure\ref{fig:framework}) in the text decoder. Specifically, during the generation process, the decoder attends to tokens already generated and all the template tokens and video tokens. The template tokens attend to themselves as well as all the video tokens. Meanwhile, following \cite{lin2022swinbert}, a Sparse Mask is used when refining the video tokens to save the computing cost. To be concrete, the Sparse Mask is learnable. Assume that the number of video tokens is $M$, and $V$ is the learnable attention mask of size $M \times M$ governing the attention among the video tokens. Then we use the sparse loss to address the redundancy among video tokens, which can be represented as:
\begin{equation}
    \setlength{\abovedisplayskip}{3pt}
    \setlength{\belowdisplayskip}{3pt}
    \mathcal{L}_{SPARSE}=\lambda\times\sum_i^M\sum_j^M |V_{i,j}|,
\end{equation} where $\lambda$ represents the regularization hyperparameter and $V_{i,j}$ represents the activation values of the attention mask.

\begin{table*}[]
\begin{center} 
\caption{\textbf{Comparison with previous video captioning methods on two benchmarks for NAE task.} Except for the NAE metric (mAP), we also specifically compare the accuracy of professional information in the generated text using AQA and Action Classification metrics. The scores and actions are extracted from the generated text. We also utilize Video Captioning metrics to assess the quality of the generated text.} 
\vspace{-5pt}
\label{tab:maintable}
\small \renewcommand{\arraystretch}{0.8}
\setlength{\tabcolsep}{1.97mm}{ 
\begin{tabular}{l|c|cc|ccc|c|c|cc|ccc|c}
\toprule
\multirow{4}{*}{Method} & \multicolumn{7}{c|}{MTL-NAE} & \multicolumn{7}{c}{FineGym-NAE}\\
\cmidrule(lr){2-8} \cmidrule(lr){9-15}
 & NAE & \multicolumn{2}{c|}{AQA} & \multicolumn{3}{c|}{Captioning} & Action 
 & NAE & \multicolumn{2}{c|}{AQA} & \multicolumn{3}{c|}{Captioning} & Action\\
\cmidrule(lr){2-8} \cmidrule(lr){9-15} & mAP & $\rho\uparrow$ & R-$\ell_2\downarrow$ & B4 & M & C & Acc & mAP & $\rho\uparrow$ & R-$\ell_2\downarrow$ & B4 & M & C & Acc  \\ 
  \midrule
C3D-AVG\cite{parmar2019and}  & 0.157 & 0.843& 1.032& 16.4 & 18.6 & 13.6 & 0.89& 0.051 & 0.606& 3.76& 10.0 & 9.8  & 11.6& 0.80\\
MSCADC\cite{parmar2019and}   & 0.074 & 0.797& 1.601& 16.7 & 18.4 & 13.3 & 0.84& 0.025 & 0.583& 4.42& 10.2 & 10.3 & 12.6& 0.76\\
UniVL\cite{luo2020univl}    & 0.166 & 0.836& 1.086& 16.4 & 18.3 & 13.6 & 0.87& 0.057 & 0.604& 3.81& 11.0 & 10.7 & 13.3& 0.79\\
VLCap\cite{yamazaki2022vlcap} & 0.197 & 0.851& 0.867& 19.8 & 18.7 & 13.9 & 0.90& 0.086 & 0.627& 3.07& 13.6 & 12.1 & 13.5& 0.81\\
VLTinT\cite{yamazaki2022vltint}   & 0.214 & 0.868& 0.820& 22.5 & 19.9 & 14.4 & 0.90& 0.094 & 0.640& 2.33& 15.4 & 17.6 & 14.2& 0.84\\
SwinBert\cite{lin2022swinbert} & 0.261 & 0.881& 0.706& 40.2 & 26.4 & 16.2 & 0.92& 0.118 & 0.656& 2.13& 27.4 & 21.0 & 17.8& 0.85\\
\midrule
\rowcolor{gray!20}\textbf{Ours} & \textbf{0.383} & \textbf{0.943} & \textbf{0.340} & \textbf{42.2} & \textbf{28.2} & \textbf{20.5} & \textbf{0.97}& \textbf{0.162} & \textbf{0.749}& \textbf{1.55}& \textbf{28.9} & \textbf{23.7} & \textbf{20.7}& \textbf{0.93}\\
\bottomrule
\end{tabular}}
\end{center}
\vspace{-15pt}
\end{table*}

\section{Experiment}

\subsection{Experimental Setup}
\paragraph{Evaluation Metrics.} 

We want our model to accurately provide evaluation information (such as score and action) and rich descriptions (as in Video Captioning), intuitively.

To simultaneously consider the performance of score prediction, action prediction, and text generation, we propose to measure the mean Average Precision (AP) across a range of thresholds for evaluation metrics in AQA \cite{yu2021group}, Action Classification, and Video Captioning \cite{vedantam2015cider}. For AQA we use intersection over relative $\ell_2$-distance (R-$\ell_2$) thresholds 0.003, 0.005, 0.010, 0.015, 0.020. For captioning we use CIDEr score thresholds .05, .10, .15, .20, .25. For Action Classification we use the average Acc of the classification results of all actions in a sentence with thresholds 0.25, 0.50, 0.75, 1. We adopt CIDEr since the idea that good captions should not only be similar to the reference captions in terms of word choice and grammar but also in terms of meaning and content \cite{vedantam2015cider}. We calculate the average precision for all the possible combinations of thresholds and report the average of APs. Thus, the mAP values range from 0 to 1.

In our experiments, we also use mainstream metrics for traditional tasks to compare with our task and method. For AQA, we use Spearman’s rank correlation ($\rho$) \cite{tang2020uncertainty, xu2022finediving, yu2021group,shiyi2023logo} and R-$\ell_2$ to measure the rank correlation and relative distance (following previous works, we multiply R-$\ell_2$ by 100). And for Captioning, we also use BLEU\cite{papineni2002bleu}, METEOR\cite{banerjee2005meteor} to assess the language generation capability of the models. Among these metrics, the lower the R-$\ell_2$, the better the performance. While for all other metrics, the higher, the better.

\vspace{-12pt}
\paragraph{Implementation Details.}
We conduct experiments on our re-annotated datasets, referred to as \textbf{MTL-NAE} and \textbf{FineGym-NAE} respectively. Following the original split for MTL-AQA in \cite{parmar2019and}, we divide both datasets into a training set and a test set at a ratio of 3:1 (5295 for training and 1765 for testing in MTL-NAE, 4665 for training and 1560 for testing in FineGym-NAE). We use the Video Swin Transformer \cite{liu2022video} pre-trained on the Kinetics600 dataset and CLIP \cite{radford2021learning} model with pre-trained weights provided by Huggingface. Other components are randomly initialized. We employ the AdamW optimizer and include a learning rate warm-up during the first 10\% of training steps followed by linear decay. More details can be found in the supplementary materials.

\subsection{Results on Narrative Action Evaluation}
As shown in Table \ref{tab:maintable}, we conduct NAE experiments on two re-annotated datasets. The results show that the mAP for the NAE task of our method is significantly better than previous methods (achieving improvements of 46.7\% and 37.3\% on MTL-NAE and FineGym-NAE respectively). To demonstrate more precisely that our method can generate sentences that balance evaluation accuracy and linguistic richness, we extract score and action information from the generated sentences to calculate evaluation metrics for the AQA and Action Classification tasks. We also calculate metrics for the Captioning task. It can be seen that our method outperforms previous methods in these three subtasks, especially in predicting professional evaluation information (scores and action types). The main reason is that previous methods ignore the accuracy of professional information when generating text. Due to the flexibility of text generation, professional evaluation information often only accounts for a small proportion of sentences. Therefore, traditional methods cannot accurately provide professional evaluations during text generation. Our method, however, can guide multi-task interactions and introduce professional information into the text generation to generate accurate evaluation information.

\subsection{Ablation Study}

\begin{table}[t]
\caption{\textbf{Comparison with existing multi-task learning methods on MTL-NAE.} We change the backbone in \cite{parmar2019and} to Video Swin Transformer\cite{liu2022video} for fair comparisons. For AQA and Action, methods with $*$ use scores and actions from their regression and classification heads. Ours uses scores and actions from the generated sentences except for the AQA-only case in the first line.}
\vspace{-5pt}
\label{tab:multitaskablation}
\centering
\fontsize{8}{14}\selectfont \renewcommand{\arraystretch}{0.6}
\setlength{\tabcolsep}{0.75mm}{ 
\begin{tabular}{l|c|c|cc|ccc|c}
\toprule
\multirow{2}{*}{Method} & \multirow{2}{*}{Tasks} & NAE & \multicolumn{2}{c|}{AQA} & \multicolumn{3}{c|}{Captioning} & Action\\ 
\cmidrule(lr){3-9} & & mAP & $\rho$ $\uparrow$ & R-$\ell_2$ $\downarrow$ & B4 & M & C & Acc\\ 
\midrule
\multirow{4}{*}{AVG*} & AQA & - & 0.903 & 0.512 & - & - & - & -\\
& Cap & 0.149 & -& -& 16.5 & 18.6 & 13.7 & -\\
& Cap+AQA & 0.151 & 0.910 & 0.472 & 16.2 & 18.4 & 13.6 & -\\
& Cap+AQA+Cls & 0.157 & 0.912 & 0.542 & 16.4 & 18.6 & 13.7 & 0.97\\
\midrule
\multirow{4}{*}{MSC*} & AQA & - & 0.863 & 0.840 & - & - & - & -\\
& Cap & 0.071 & -& -& 17.0 & 18.6 & 13.5 & -\\
& Cap+AQA & 0.072 & 0.857 & 0.847 & 16.5 & 18.3 & 13.4 & -\\
& Cap+AQA+Cls & 0.074 & 0.860 & 0.842 & 16.7 & 1804 & 13.3 & 0.84\\
\midrule 
\rowcolor{gray!20}& AQA & - & 0.909 & 0.633 & - & - & - & -\\
\rowcolor{gray!20}& Cap & 0.341 & 0.897 & 0.569 & 40.9 & 27.7 & 17.8 & 0.92\\
\rowcolor{gray!20}& Cap+AQA & 0.379 & 0.940 & 0.346 & 41.7 & 27.6 & 19.5 & 0.93\\
\rowcolor{gray!20}\multirow{-4}{*}{\textbf{Ours}}& Cap+AQA+Cls & \textbf{0.383} & \textbf{0.943} & \textbf{0.340} & \textbf{42.2} & \textbf{28.2} & \textbf{20.5} & \textbf{0.97}\\
\bottomrule
\end{tabular}}
\vspace{-5pt}
\end{table}

\textbf{Effectiveness of Our Multi-Task Learning Paradigm.} To verify the superiority of our method compared to the traditional multi-task learning paradigm, we conduct comparisons with the multi-task learning method, C3D-AVG, and MSCADC, from \cite{parmar2019and}. These methods jointly train a backbone with independent branches. Our approach differs from them by coupling the tasks closer with prompt-guided multimodal interaction. We compare the strategies on the MTL-NAE dataset with the visual backbone in \cite{parmar2019and} changed to Video Swin Transformer \cite{liu2022video} for a fair comparison. As shown in Table \ref{tab:multitaskablation}, our multi-task learning method brings significant performance gain to single-task learning. Specifically, compared to the single AQA training mode, our method shows a significant performance improvement, with R-l2 decreasing by 46.3\%. Compared to the single Captioning training mode, all metrics have improved, with CIDEr increasing by 15.2\%. However, the method in \cite{parmar2019and} improves little and even brings degradation to their single-task method. These results demonstrate the effectiveness of our proposed prompt-guided multimodal interaction multi-task learning.



\begin{table}[]
\begin{center} 
\caption{\textbf{Ablation study of different components.} \textit{CAT.} means Context-Aware Transformer, and \textit{SAT.} is Score-Aware Transformer.}
\vspace{-5pt}
\label{tab:frameworkablation}
\fontsize{8}{14}\selectfont \renewcommand{\arraystretch}{0.67}
\setlength{\tabcolsep}{1.41mm}
\begin{tabular}{l|c|cc|ccc|c}
\toprule

\multirow{2}{*}{w/o Module} & NAE & \multicolumn{2}{c|}{AQA} & \multicolumn{3}{c|}{Captioning} & Action\\ 
\cmidrule(lr){2-8} & mAP & $\rho$ $\uparrow$ & R-$\ell_2$ $\downarrow$ & B4 & M & C & Acc\\ 

\midrule
CAT. & 0.352 & 0.919& 0.428& 41.46 & 27.37 & 19.08 & 0.95\\
SAT. & 0.347 & 0.886& 0.589& 41.73 & 27.69 & 19.68 & 0.95\\
\midrule
\rowcolor{gray!20}\textbf{Ours} & \textbf{0.383} & \textbf{0.943}& \textbf{0.340}& \textbf{42.23} & \textbf{28.22} & \textbf{20.54} & \textbf{0.97}\\
\bottomrule
\end{tabular}
\end{center}
\vspace{-20pt}
\end{table}

\vspace{-8pt}
\paragraph{Effect of Different Framework Components.} We then investigate the effectiveness of various components in our framework. As shown in Table \ref{tab:frameworkablation}, we observe the following two facts. Firstly, without the Context-Aware Transformer, the model fails to integrate textual score information with contextual information in videos. Consequently, removing the Context-Aware Transformer results in a 25.9\% increase in R-$\ell_2$ and a drop from 20.54 to 19.08 in CIDEr, ultimately leading to an 8.8\% decrease in mAP. Secondly, if video information is not guided to perceive language modality scores by the Score-Aware Transformer, the integration of video information and score information becomes impossible, leading to a significantly worse performance in AQA metrics where $\rho$ drops from 0.943 to 0.886 and the R-$\ell_2$ increases by 73.2\%, ultimately resulting in a 9.4\% decrease in the mAP.

\begin{table}[]
\begin{center}
\caption{\textbf{Ablation study of loss functions on MTL-NAE dataset.}}
\vspace{-5pt}
\fontsize{8}{14}\selectfont \renewcommand{\arraystretch}{0.67}
\setlength{\tabcolsep}{1.58mm}
\begin{tabular}{l|c|cc|ccc|c}
\toprule
\multirow{2}{*}{w/o Loss} & NAE & \multicolumn{2}{c|}{AQA} & \multicolumn{3}{c|}{Captioning} & Action\\ 
\cmidrule(lr){2-8} & mAP & $\rho$ $\uparrow$ & R-$\ell_2$ $\downarrow$ & B4 & M & C & Acc\\ 
\midrule
CE & 0.308 & 0.881 & 0.652 & 41.62 & 27.32 & 18.96 & 0.97\\
MSE & 0.316 & 0.876 & 0.651 & 41.56 & 27.35 & 19.22 & 0.97\\
CE+MSE & 0.293 & 0.851 & 0.783 & 41.38 & 27.44 & 18.74 & 0.97\\
\midrule
\rowcolor{gray!20}\textbf{Ours}& \textbf{0.383} & \textbf{0.943} & \textbf{0.340} & \textbf{42.23} & \textbf{28.22} & \textbf{20.54} & \textbf{0.97}\\
\bottomrule
\end{tabular}
\label{tab:lossablation}
\end{center}
\vspace{-15pt}
\end{table}

\vspace{-8pt}
\paragraph{Effect of Different Loss Functions.} We adopt two loss functions to incorporate the score information, namely mean squared error loss (MSE), which directly regulates predicting scores, and cross-entropy loss (CE), which plays a vital role in regulating the Score-Aware Transformer. We conduct experiments by removing one or both of them, and the results are shown in Table \ref{tab:lossablation}. We observe that no matter which loss function is removed, it will lead to a significant decrease in the final AQA metrics, and then result in a significant decrease in the NAE metric mAP. This result proves the effectiveness of both loss functions. The NAE task has a high sensitivity to the accuracy of professional information. Even though these two loss functions related to score prediction have little impact on text generation performance, they can supervise the model to generate more accurate evaluation information, thus improving the performance of NAE task.

\begin{table}[]
\begin{center} 
\caption{\textbf{Ablation study of different kinds of templates on MTL-NAE dataset.} \textit{w/o Lr.} means the template without learnable tokens. \textit{All Lr.} denotes the template composed entirely of learnable tokens.}
\label{tab:templateablation}
\fontsize{8}{14}\selectfont \renewcommand{\arraystretch}{0.67}
\setlength{\tabcolsep}{1.58mm}
\begin{tabular}{l|c|cc|ccc|c}
\toprule
\multirow{2}{*}{Template} & NAE & \multicolumn{2}{c|}{AQA} & \multicolumn{3}{c|}{Captioning} & Action\\ 
\cmidrule(lr){2-8} & mAP & $\rho$ $\uparrow$ & R-$\ell_2$ $\downarrow$ & B4 & M & C & Acc\\ 
\midrule
w/o Lr.& 0.364 & 0.935& 0.366& 41.02 & 27.77 & 19.54 & 0.95\\
All Lr. & 0.258 & 0.898& 0.605& 36.90 & 25.99 & 14.21 & 0.90\\
\midrule
\rowcolor{gray!20}\textbf{Ours} & \textbf{0.383} & \textbf{0.943}& \textbf{0.340}& \textbf{42.23} & \textbf{28.22} & \textbf{20.54} & \textbf{0.97}\\
\bottomrule
\end{tabular}
\end{center}
\vspace{-20pt}
\end{table}

\begin{figure*}[t]
    \centering
    \includegraphics[width=\linewidth]{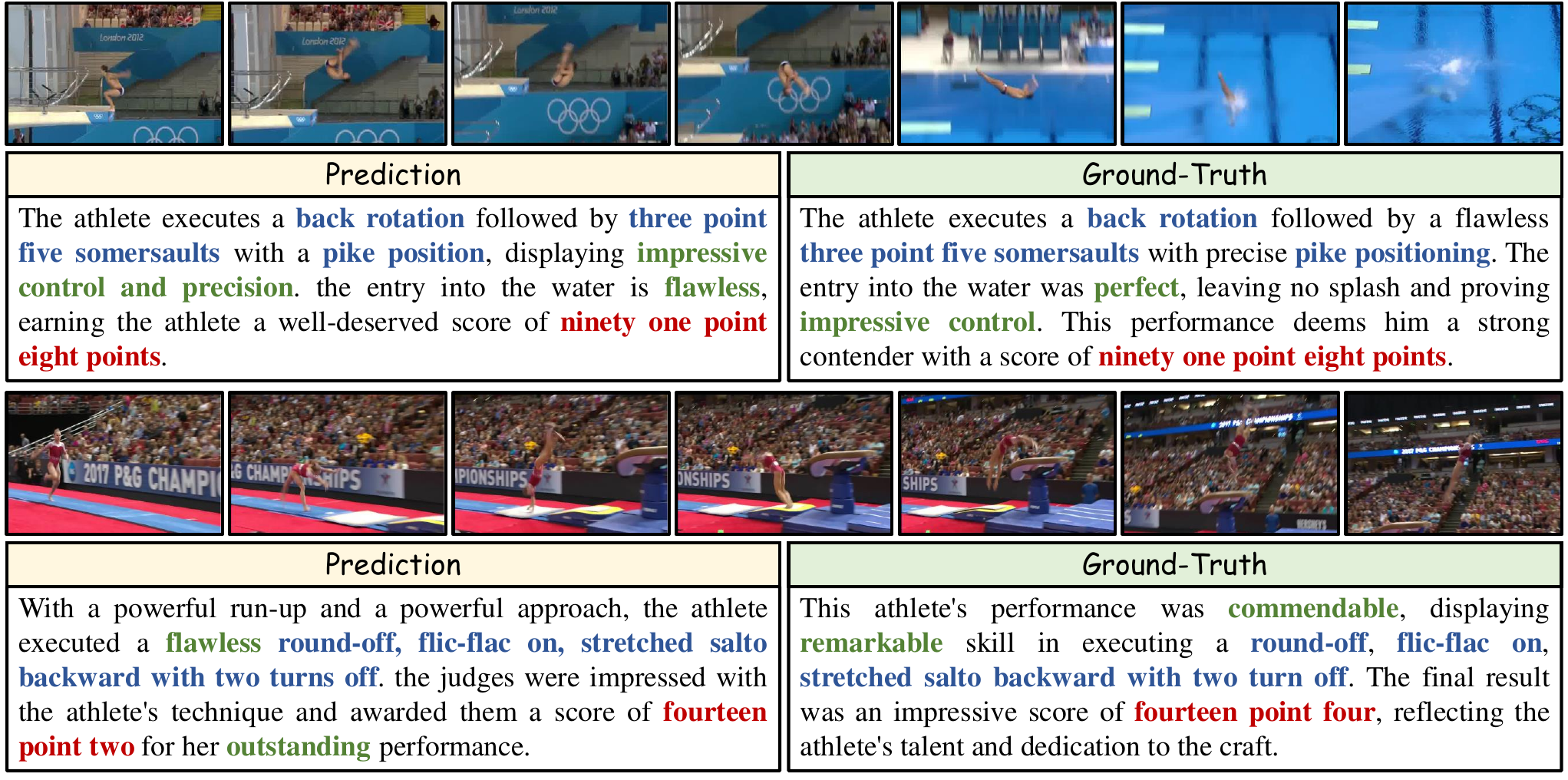}
    \vspace{-20pt}
    \caption{\textbf{Qualitative results.} Our model can generate detailed narrations including \textcolor[RGB]{192,0,0}{\textbf{\underline{\textit{scores}}}}, \textcolor[RGB]{47,85,151}{\textbf{\underline{\textit{actions}}}}, and \textcolor[RGB]{109,149,82}{\textbf{\underline{\textit{qualitative evaluations}}}} to describe and evaluate the actions comprehensively. Notably, the model can analyze the quality of actions by pointing out the details of the execution.}
    \label{fig:case}
    \vspace{-12pt}
\end{figure*}
\vspace{-8pt}
\paragraph{Effect of Learnable Template.} To demonstrate the effectiveness of our template design, we compared the performance of models under three different templates in Table \ref{tab:templateablation}. The three templates are \textit{w/o Lr.} (only contains fixed evaluation information tokens and without learnable tokens), \textit{All Lr.} (only contains learnable tokens), and \textit{ours} (contains both evaluation tokens and learnable tokens). It can be seen that \textit{w/o Lr.} introduces evaluation information into the text generation process, thus significantly outperforms \textit{All Lr.}, which includes no evaluation information. While its performance is worse than the template inserted with the learnable tokens.

\begin{table}[]
\begin{center} 
\vspace{-5pt}
\caption{\textbf{Comparison with state-of-the-art AQA methods on two benchmarks.} Although the NAE task aims to generate comprehensive natural language assessments, our framework outperforms all of the methods that use a single video as input on both datasets.}
\fontsize{8}{14}\selectfont \renewcommand{\arraystretch}{0.63}
\setlength{\tabcolsep}{3.2mm}
\label{tab:aqa}
\begin{tabular}{l|cc|cc}
\toprule
\multirow{2}{*}{Method} & \multicolumn{2}{c|}{MTL-NAE} & \multicolumn{2}{c}{FineGym-NAE} \\ 
\cmidrule(lr){2-3} \cmidrule(lr){4-5} & $\rho$ $\uparrow$ & R-$\ell_2$ $\downarrow$ & $\rho$ $\uparrow$ & R-$\ell_2$ $\downarrow$\\ 
  \midrule
  \multicolumn{5}{l}{\emph{Methods with single input video}} \\
  \midrule
C3D-LSTM\cite{parmar2017learning} & 0.849 & - & 0.641 & - \\
C3D-AVG-MTL\cite{parmar2019and} & 0.904 & - & 0.701 & - \\
USDL\cite{tang2020uncertainty} & 0.923 & 0.468 & 0.726 & 1.82 \\
MUSDL\cite{tang2020uncertainty} & 0.927 & 0.541 & 0.729 & 1.78 \\
\rowcolor{gray!20}\textbf{Ours} & \textbf{0.943} & \textbf{0.340} & \textbf{0.749} & \textbf{1.55} \\
  \midrule
  \multicolumn{5}{l}{\emph{Methods with several input videos}} \\
  \midrule
CoRe\cite{yu2021group} & 0.951 & 0.260 & 0.754 & 1.34 \\
TPT\cite{bai2022action} & 0.961 & 0.238 & 0.764 & 1.27 \\
\bottomrule
\end{tabular}
\end{center}
\vspace{-29pt}
\end{table}

\subsection{Analysis on Action Quality Assessment}
While the primary objective of the NAE task is to generate detailed narrations in language form, we also compare the performance of the score prediction with existing state-of-the-art methods in traditional score-based AQA \cite{tang2020uncertainty, yu2021group, parmar2019and, bai2022action}. The results are shown in Table \ref{tab:aqa}. Notably, our model attains the result of 0.943 on Spearman's rank correlation and 0.340 on relative $\ell_2$-distance on MTL-NAE, surpassing all the methods that use a single input video. Besides, our model achieves comparable performance with recently proposed methods that require additional exemplar videos \cite{yu2021group,bai2022action}, which predict score differences by comparing multiple input videos. Such a paradigm is suitable for comparative tasks like predicting score difference, but not for our NAE task that needs to focus on the information of a single video since the input of multiple videos may introduce noises. These experimental results prove that our approach can predict the score accurately, even only using a single video as the input.
\vspace{-20pt}
\subsection{Qualitative Results}
\vspace{-5pt}
In Figure \ref{fig:case}, we display qualitative examples of our model. We observe that our model is capable of generating detailed narrations that describe the corresponding action categories and scores. Notably, our model can assess and analyze the quality of actions, as indicated by phrases such as ``entry...flawless'' and ``impressive control'', highlighting commendable execution and areas where improvements can be made. Besides, our predicted action categories and scores are accurate and basically the same as the ground truth. More qualitative results can be found in supplementary materials.

\vspace{-21pt}
\section{Conclusion}
\vspace{-6pt}
In this paper, we have introduced the Narrative Action Evaluation (NAE) task, which aims to generate professional commentary to assess action executions maintaining both narrative flexibility and evaluation rigor. To address this task, we have proposed a Prompt-Guided Multimodal Interaction multi-task learning framework, which interacts and integrates different tasks and information from different modalities, thereby achieving performance improvement in the NAE task and multiple subtasks. To facilitate further research in this field, we have re-annotated the MTL-AQA and FineGym datasets and established benchmarks for the NAE problem. Extensive experimental results have demonstrated the power and efficiency of our model with respect to baselines based on the previous state-of-the-art methods.
\vspace{-16pt}
\paragraph{Acknowledgments.} This work was supported in part by the National Natural Science Foundation of China under Grant 62125603, Grant 62321005, and Grant 62336004, and in part by CCF-Tencent Rhino-Bird Open Research Fund.
{
    \small
    \bibliographystyle{ieeenat_fullname}
    \bibliography{main}
}


\end{document}